\documentclass{article}
\usepackage{spconf,amsmath,graphicx,hyperref}
\usepackage{times}
\usepackage{latexsym}
\usepackage{graphicx}
\usepackage{amsmath}
\usepackage{amssymb}
\usepackage{multirow}
\usepackage{graphicx}
\usepackage{booktabs}
\usepackage{subcaption} 
\usepackage{float} 
\usepackage{xcolor} 
\usepackage{adjustbox}
\usepackage{booktabs} 
\usepackage{booktabs}
\usepackage{wrapfig}   
\usepackage{xcolor}    
\usepackage{caption} 
\definecolor{softpeach}{RGB}{251, 233, 221}
\definecolor{softblue}{RGB}{240, 248, 255} 
\usepackage[table]{xcolor} 
\definecolor{mycolor}{HTML}{BAD8F2}
\usepackage[T1]{fontenc}
\usepackage{enumitem} 

\usepackage[utf8]{inputenc}

\usepackage{microtype}

\usepackage{inconsolata}
\DeclareRobustCommand{\OurMethod}{\textsc{HORSE}}

\usepackage{graphicx}
\usepackage{amsmath} 
\usepackage{amssymb} 

\title{Hierarchical Orthogonal Residual Spread for Precise Massive Editing in Large Language Models}
\name{Xiaojie Gu$^{1}$\textsuperscript{*}, Guangxu Chen$^{2}$\textsuperscript{*}, Yuheng Yang$^{1}$, Jingxin Han$^{3}$, Andi Zhang$^{4}$
}
  
  \address{$^{1}$ Independent Researcher \qquad $^{2}$UESTC 
    \qquad $^{3}$Shanghai University 
  \qquad $^{4}$University of Manchester
  }

\begin{document}
%

\maketitle

\begingroup
\renewcommand\thefootnote{\fnsymbol{footnote}} 
\footnotetext[1]{Equal contribution. Correspondence to: \href{mailto:peettherapynoys@gmail.com}{peettherapynoys@gmail.com}.}
\endgroup

\begin{abstract}
Large language models (LLMs) exhibit exceptional performance across various domains, yet they face critical safety concerns. Model editing has emerged as an effective approach to mitigate these issues.
Existing model editing methods often focus on optimizing an information matrix that blends new and old knowledge. While effective, these approaches can be computationally expensive and may cause conflicts.
In contrast, we shift our attention to \textbf{H}ierarchical \textbf{O}rthogonal \textbf{R}esidual \textbf{S}pr\textbf{E}ad of the information matrix, which reduces noisy gradients and enables more stable edits from a different perspective.
We demonstrate the effectiveness of our method \OurMethod\ through a clear theoretical comparison with several popular methods and extensive experiments conducted on two datasets across multiple LLMs. 
The results show that \OurMethod\ maintains precise massive editing across diverse scenarios. 
The code is available at \url{https://github.com/XiaojieGu/HORSE}

\end{abstract}


\section{Introduction}

The world is evolving, and humans acquire knowledge through long term learning. Large language models (LLMs)\cite{gptj,llama2,mistral} demonstrate strong capabilities across domains, fueling progress toward artificial general intelligence (AGI). However, during pre-training they may memorize sensitive information\cite{halsurvey1}, leading to hallucinations and security risks~\cite{pmo}.

To address these issues, model editing has emerged as a promising direction~\cite{easyedit}, enabling targeted updates (\textit{Efficacy} \& \textit{Generalization}) without retraining while preserving the original knowledge (\textit{Specificity}).
Most editing methods~\cite{rome,memit,emmet} follow a \textit{locate-then-edit} paradigm. 
They use causal tracing to identify layers responsible for storing a fact, and then optimize a residual matrix that integrates new knowledge with existing parameters. However, this process often requires substantial computation and can introduce conflicts between new and old knowledge~\cite{easyedit}.
Adding \textit{external memory}\cite{wise} to store introduced knowledge can alleviate conflicts between old and new information, but continual updates often reduce retrieval accuracy and increase the risk of catastrophic forgetting\cite{alphaedit}.
\textit{Meta-learning-based} methods~\cite{mend,malmen} reduce cost by using hypernetworks, but they may introduce instability at scale~\cite{ultraedit}.

We introduce \OurMethod\ to tackle the above challenges.
First, we operate at the token level for fine-grained control, enhancing stability under high-dimensional edits. 
Then, instead of treating the residual as a monolithic update, \OurMethod\ performs a \emph{hierarchical orthogonal spread} across layers and adapts layer-wise weights in real time, reducing reliance on a single layer and mitigating overfitting and cross-knowledge interference~\cite{memit}. Unlike linear or fixed schedules~\cite{memit,malmen}, this adaptive orthogonal scheme better matches how knowledge is distributed in transformers. \OurMethod\ also trains the hypernetwork on residual information rather than full weight updates, improving alignment with editing goals and robustness.

On the experimental front, we evaluate \OurMethod\ on two popular datasets with GPT, LLaMA, and Mistral. \OurMethod\ achieves state-of-the-art performance across most editing scenarios while maintaining the fastest editing speed. 
The results indicate that \OurMethod\ delivers strong stability for massive and precise editing. Moreover, comprehensive ablation studies and analyses of the post-edited models confirm not only the effectiveness and efficiency of the proposed approach but also its minimal impact on the original model’s capabilities.


In this work, we propose \OurMethod\ to address key challenges in model editing through hierarchical orthogonal residual spread of the information matrix, thereby improving stability, mitigating knowledge conflicts, and enabling adaptive updates in real time. Beyond traditional optimization of the information matrix itself, we highlight the less-explored perspective of residual propagation across layers as a promising direction for future research. Extensive experiments show that \OurMethod\ achieves the \textbf{fastest} editing speed to date and delivers consistent performance improvements, with an average gain of \textbf{+6.26\%} over prior methods. In particular, it achieves a notable \textbf{+10.12\%} improvement in Specificity, effectively reducing interference with existing knowledge while preserving factual alignment and generalization.

\begin{figure*}[t]
    \centering
    \includegraphics[scale=0.45]{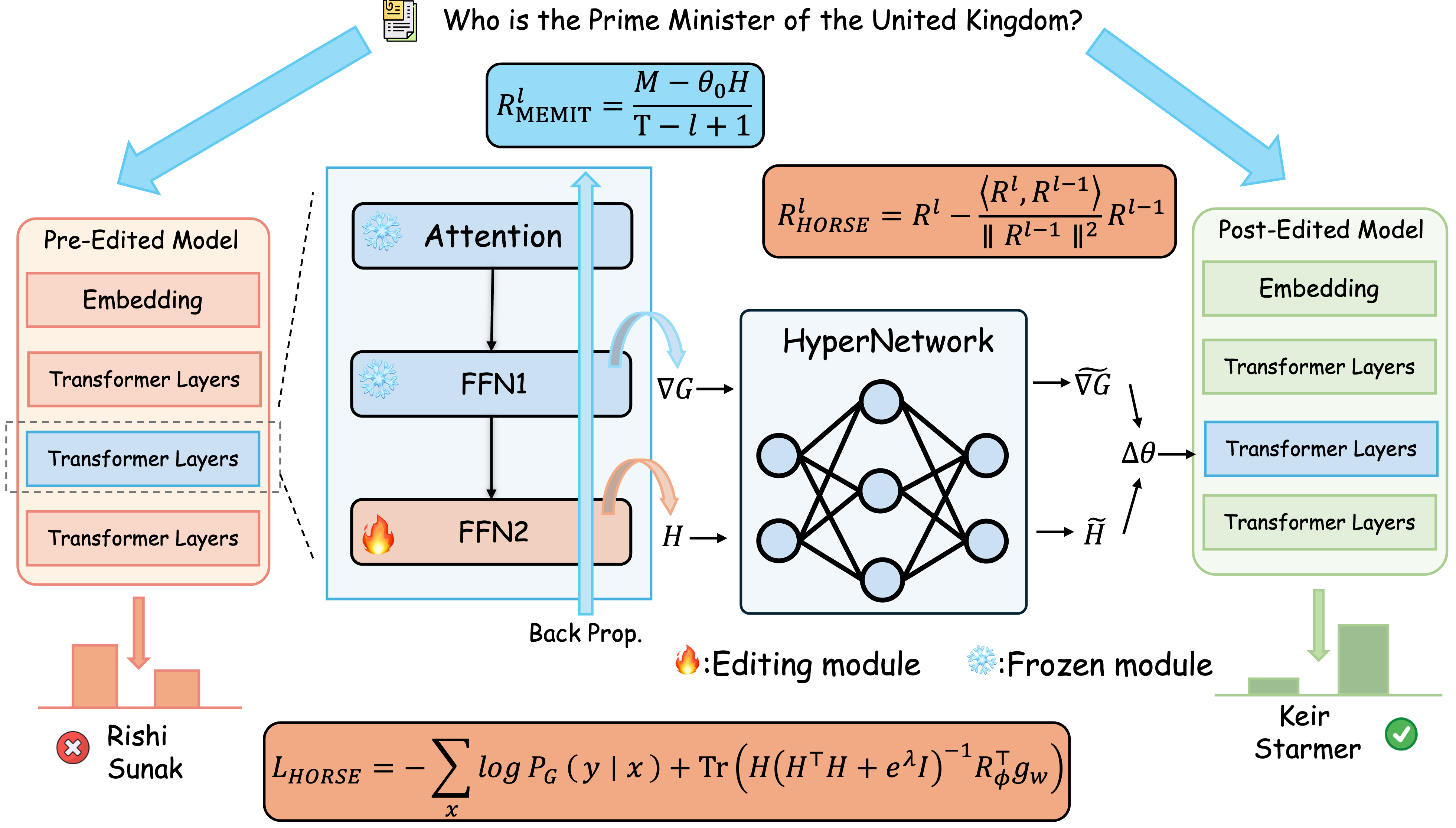} 
    \caption{ Overall structure of \OurMethod. \(\Delta \theta\) denotes the weight update, \(R\) denotes the residual matrix, which is the key variable used to generate \(\Delta \theta\),
    and \(L\) denotes the loss for training the hypernetwork.
    Details of the HyperNetwork are provided in Section~\ref{HyperNetwork}.}
    
    \label{pic_main}
\end{figure*}

\section{Method}

In this section, we first introduce the task of model editing and then elaborate on the editing process by comparing two widely used methods, MEMIT and MALMEN, which face various challenges, including instability in editing and insufficient knowledge retention.
Through this comparison, we aim to clarify our approach, \OurMethod\, and highlight its key distinctions and improvements over these methods.
The overall framework of \OurMethod\ can be found in Figure~\ref{pic_main}.

\subsection{Preliminary}

Model editing modifies the parameters of a pre-trained language model \( f_\theta \) into \( f_{\theta'} \) to integrate new knowledge.
Given an \textit{editing instance} \((x_e, y_e)\), the goal is to ensure \( f_{\theta'}(x_e) = y_e \), while \( f_\theta(x_e) \neq y_e \).
The edited model should also generalize to \textit{equivalent instances} \( x'_e \in \mathcal{E}(x_e) \) and preserve behavior on \textit{unrelated instances} \( x_u \in \mathcal{U}(x_e) \).
These three instance types correspond to the metrics \textit{Efficacy (Eff.)}, \textit{Generalization (Gen.)}, and \textit{Specificity (Spe.)}.

To realize this, editing methods typically optimize a transformation matrix \( \theta' \in \mathbb{R}^{d' \times d} \) that integrates new knowledge while retaining the original~\cite{memit,pmet,emmet}. 
Given a batch of \( n \) instances, we extract two representation from the editable module (We strictly follow MALMEN, selecting the second MLP block within each of the last six layers.):  
a \textit{forward hook} records the hidden state \( h_i \in \mathbb{R}^d \) at the label position of each editing instances, and a \textit{backward hook} collects the gradient \( \nabla G \in \mathbb{R}^{d'} \) from the supervised loss.  
The loss combines cross-entropy over editing, equivalent, and unrelated instances, ensuring precise updates while preserving prior knowledge~\cite{malmen}. 
Concatenating hidden states yields \( H \in \mathbb{R}^{d \times n} \). With a residual matrix \( R \in \mathbb{R}^{d' \times n} \) encoding the updates, the objective becomes:
\begin{equation}
\Delta \theta^* = \arg\min_{\Delta\theta}  \| \Delta\theta H - R \|_F^2 + \lambda \| \Delta\theta \|_F^2,
\end{equation}
where \( \theta_0 \) is the pre-trained weight and \( \lambda \) controls the balance between preservation and update.  
The closed-form solution is:
\begin{equation}
\Delta \theta^* = R H^T (H H^T + \lambda I)^{-1},
\label{eq2}
\end{equation}
where \( \lambda I \) is a regularization term that stabilizes matrix inversion. 
In implementation, \(\lambda\) is learned for each editable layer and exponentiated to ensure it remains positive, 
with the initial value set to 1.
The final edited parameters are then obtained as \( \theta' = \theta_0 + \Delta \theta \).

\begin{table*}[!htb]
\scriptsize
\centering
\setlength{\tabcolsep}{3pt}
\renewcommand{\arraystretch}{1.1}
\resizebox{\textwidth}{!}{%
\begin{tabular}{l|ccc|ccc|ccc|ccc|ccc|ccc}
\toprule[1.5pt]
\multirow{3}{*}{\textbf{Method}} 
& \multicolumn{9}{c|}{\textbf{zsRE}} 
& \multicolumn{9}{c}{\textbf{CounterFact}} \\
\cmidrule(lr){2-10} \cmidrule(lr){11-19}
& \multicolumn{3}{c|}{\textbf{GPT-J }} 
& \multicolumn{3}{c|}{\textbf{LLaMA2 }} 
& \multicolumn{3}{c|}{\textbf{Mistral }} 
& \multicolumn{3}{c|}{\textbf{GPT-J }} 
& \multicolumn{3}{c|}{\textbf{LLaMA2 }} 
& \multicolumn{3}{c}{\textbf{Mistral }} \\
\cmidrule(lr){2-4} \cmidrule(lr){5-7} \cmidrule(lr){8-10} 
\cmidrule(lr){11-13} \cmidrule(lr){14-16} \cmidrule(lr){17-19}
& \textbf{Eff.} & \textbf{Gen.} & \textbf{Spe.} 
& \textbf{Eff.} & \textbf{Gen.} & \textbf{Spe.} 
& \textbf{Eff.} & \textbf{Gen.} & \textbf{Spe.} 
& \textbf{Eff.} & \textbf{Gen.} & \textbf{Spe.} 
& \textbf{Eff.} & \textbf{Gen.} & \textbf{Spe.} 
& \textbf{Eff.} & \textbf{Gen.} & \textbf{Spe.}  \\
\midrule
FT      
& 23.40 & 21.93 & 16.42 & 32.74 & 32.40 & 44.45 & 35.36 & 34.89 & \underline{46.61}
& 11.82 & 6.84  & 26.79 & 17.52 & 13.98 & 20.35 & 13.89 & 10.97 & 34.50 \\
LoRA    
& 34.28 & 32.91 & \underline{27.01} & 50.58 & 48.05 & 44.77 & 53.80 & 47.48 & 38.59
& 11.93 & 5.47  & 30.11 & 30.09 & 21.76 & \underline{49.36} & 32.27 & 21.95 & 41.57 \\
MEMIT   
& 86.17 & 67.22 & 25.95 & 59.56 & 54.06 & 26.99 & 62.21 & 55.39 & 26.16
& \underline{95.54} & \textbf{36.49} & 11.63 & 53.17 & 34.16 & 13.77 & 63.64 & 34.01 & 10.63 \\
MEND    
& 0.16  & 0.18  & 0.02  & 2.56  & 2.56  & 3.46  & 1.92  & 1.92  & 1.89
& 0.00  & 0.00  & 0.00  & 0.00  & 0.00  & 0.00  & 0.00  & 0.00  & 0.00  \\
PMET    
& 22.64 & 21.16 & 24.04 & 3.17  & 2.68  & 0.07  & 0.02  & 0.03  & 0.12
& 6.61  & 2.73  & 14.89 & 16.01 & 12.43 & 36.69 & 15.43 & 10.98 & 26.52 \\
EMMET   
& 76.35 & 63.48 & 26.25 & 34.85 & 32.65 & 17.22 & 41.05 & 38.14 & 18.64
& 86.37 & 29.75 & 8.70  & \underline{59.62} & 32.61 & 15.30 & 71.78 & 33.01 & 14.89 \\
MALMEN  
& \underline{98.75} & \textbf{88.52} & 24.49 & \underline{90.14} & \underline{82.86} & \underline{45.82} & \textbf{93.02} & \underline{85.26} & 45.85
& 62.42 & 22.34 & \underline{39.13} & 46.03 & 26.18 & 38.01 & \textbf{94.85} & \underline{50.54} & \underline{63.55} \\
\rowcolor{softpeach}
\textbf{\OurMethod}
& \textbf{98.91} & \underline{88.43} & \textbf{29.37} & \textbf{90.25} & \textbf{86.27} & \textbf{60.24} & \underline{92.32} & \textbf{87.28} & \textbf{53.28}
& \textbf{96.12} & \underline{35.23} & \textbf{53.00} & \textbf{83.46} & \textbf{57.83} & \textbf{72.71} & \underline{94.76} & \textbf{50.86} & \textbf{63.59} \\
\rowcolor{softblue}
$\Delta$ 
& +0.16 & -0.09 & +2.36 
& +0.11 & +3.41 & +14.42
& -0.70 & +2.02 & +6.67 
& +0.58 & -1.26 & +13.87 
& +23.84 & +23.67 & +23.35 
& -0.09 & +0.32 & +0.04 \\
\bottomrule[1.5pt]
\end{tabular}
}
\caption{
Results on zsRE and CounterFact across three models with 8,000 edits. 
\textbf{Bold} values indicate the best performance, and \underline{underlined} values indicate the second best. 
$\Delta$ indicates the performance difference between \OurMethod\ and the previous best.
}
\label{main}
\end{table*}

\subsection{Residual Matrix Optimization}
\label{method}
As shown in Eq.~\ref{eq2}, effective editing depends on obtaining the residual matrix \( R \), which captures the gap between new knowledge and existing parameters and guides updates with minimal disruption. 

In MEMIT, the residual matrix is defined as:
\begin{equation}
R_{\text{MEMIT}} = M - \theta_0 H,
\end{equation}
where \( M = H + \delta H \) represents hidden states aligned with the desired factual knowledge.  
Here, \(\delta H\) is optimized via a supervised loss for correct outputs, but directly modifying hidden states may cause excessive adjustments and conflicts~\cite{mend}.

To overcome these limitations, we follow MALMEN and employ a hypernetwork (see Section~\ref{HyperNetwork}) that takes \((H, \nabla G)\) as input and outputs transformed representations \(\tilde{H}\) and refined gradients \(\nabla \tilde{G}\).
The residual update is then formulated as: 
\begin{equation}
R_{\text{\OurMethod}} = -\eta \sum_{i=1}^{T} \sum_{j=1}^{d} H_{ij}\tilde{H}_{ij} \nabla \tilde{G},
\end{equation}
where \(\eta\) is a layer-wise learning rate. This token-level operation reduces gradient scaling costs and promotes stability.

\textbf{How to spread the residual matrix \( R \) across Transformer layers} is a crucial challenge for effective knowledge integration.  
Existing strategies are limited: MEMIT distributes updates using linear decay, which fails to capture the varying importance of hierarchical levels:
\begin{equation}
R_{\text{MEMIT}}^l = \frac{M - \theta_0 H}{T - l + 1},
\end{equation}
where \( T \) is the total number of layers.  
MALMEN applies uniform shifts across layers, ignoring their unequal contributions and often leading to under- or over-updates.  

In contrast, \OurMethod\ introduces a principled mechanism that \textit{orthogonalizes residuals across adjacent layers}, ensuring independence, avoiding redundancy, and allowing each layer to make a unique and balanced contribution.  
By subtracting the component of \( R^l \) aligned with \( R^{l-1} \), the updates remain stable while better leveraging the diverse and complementary roles of different layers:
\begin{equation}
R_{\text{\OurMethod}}^l = R^l - \frac{\langle R^l, R^{l-1} \rangle}{\| R^{l-1} \|^2} R^{l-1},
\label{equ2}
\end{equation}
where \( \langle \cdot, \cdot \rangle \) denotes the inner product between residuals of layers \( l \) and \( l-1 \), and \( \| \cdot \|^2 \) is the squared norm.  
This design enables \OurMethod\ to achieve more stable, non-redundant, and fine-grained knowledge integration than prior approaches.

\subsection{HyperNetwork Training}
\label{HyperNetwork}
The hypernetwork~\cite{hyp} is a neural network that operates independently of the pre-trained model. Following ~\cite{mend,malmen}, it is designed as a multi-layer perceptron (MLP) with low-rank matrices.
Its input consists of the features of editing instances, while its output is the projected features, aligned with the desired model adjustments.
In the entire pipeline, the hypernetwork plays a crucial role in aligning representations with the most informative editing directions in the feature space.

\textbf{How can the hypernetwork be trained to better align with editing objectives?}  
Unlike MALMEN, which aggregates gradient changes through \(\Delta H\) and introduces noise,  
we formulate optimization directly in the residual space.  
The hypernetwork is trained by backpropagating a loss that enforces alignment between its predicted residuals \(R_{\phi}\) and the accumulated gradient signals \(\mathbf{g}_w\).  
Here \(R_{\phi}\) is computed in the same way as the final update residual \(R_{\text{HORSE}}\), ensuring consistency between training and inference.  
This design enables the hypernetwork to learn update directions that minimize interference with existing knowledge while adapting efficiently to new edits:  
\begin{equation}
\begin{aligned}
L_{\text{\OurMethod}} &= -\sum_{x} \log P_G(y \mid x) \\
&\quad + \text{Tr} \left( H \left( H^\top H + e^{\lambda} I \right)^{-1} R_{\phi}^\top \mathbf{g}_w \right).
\end{aligned}
\end{equation}
The first term ensures factual alignment by maximizing the probability of correct outputs across diverse queries, while the second term regularizes updates through alignment with accumulated gradient information.  
By combining these objectives, the hypernetwork not only avoids numerical instability and overfitting to specific edits but also achieves robust and generalizable model coherence across large-scale editing scenarios.

\section{Experiments}

\subsection{Experimental Setting}

\noindent\textbf{Dataset\&Model}

\noindent 
We evaluate \OurMethod\ on two mainstream datasets: \textbf{zsRE} (question-answer pairs)~\cite{zsre}, \textbf{CounterFact} 
(counterfactual statements)~\cite{rome}.
The evaluation is conducted on three representative models: GPT-J, LLaMA-2-7B-hf, and Mistral-7B-v0.1.

\noindent\textbf{Baselines}

\noindent We compare \OurMethod\ against fine-tuning (FT) and LoRA, as well as popular massive editing methods, including MEMIT~\cite{memit}, MEND~\cite{mend}, PMET~\cite{pmet}, MALMEN~\cite{malmen}, EMMET~\cite{emmet}
We exclude AlphaEdit~\cite{alphaedit} and UltraEdit~\cite{ultraedit} as they are not designed for massive editing.

\noindent\textbf{Implementation details}

\noindent All experiments are conducted on a single A800 GPU. The hypernetwork has two MLP blocks. Learning rates are 1e-6 for zsRE, and for CounterFact: 2e-5 for GPT-J, 1e-5 for LLaMA, and 1e-6 for Mistral.

\begin{figure}[!htbp]
  \centering
  \includegraphics[width=\columnwidth]{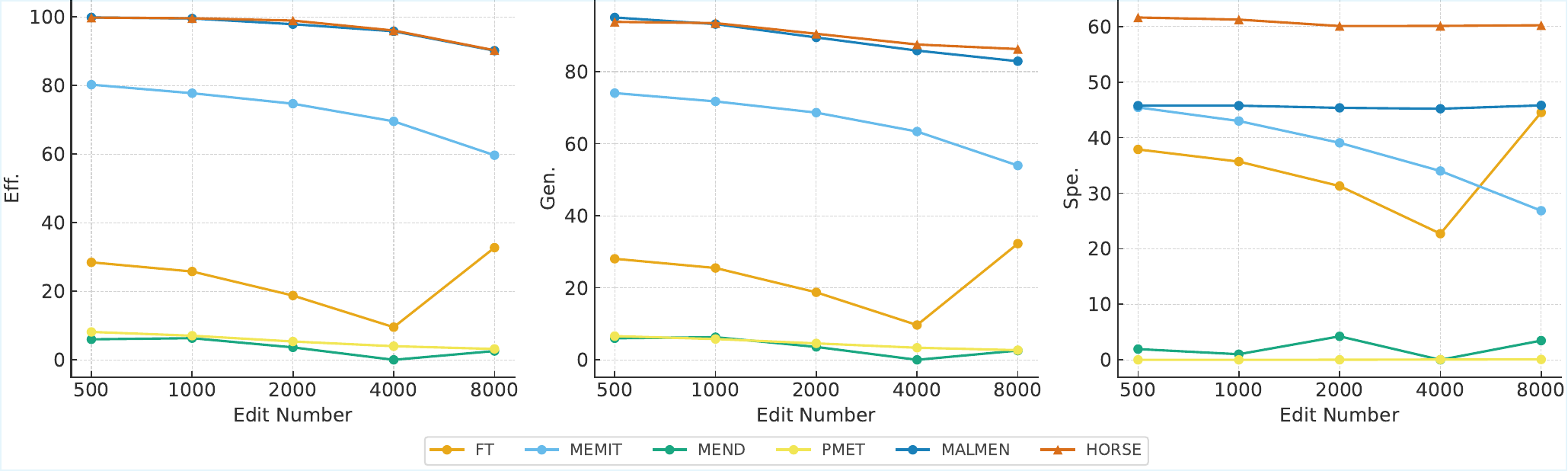} %
  \caption{Performance under different numbers of edits.}
  \label{num_batch}
\end{figure}

\subsection{Overall Results}
The results in Table~\ref{main} show that \OurMethod\ consistently achieves superior performance across diverse editing scenarios. Even with 8,000 edits, it remains stable across all three backbone models and outperforms alternatives on most key metrics. On average, it yields a 6.26\% improvement over the previous best method, including a notable 10.12\% gain in Specificity, demonstrating its ability to reduce interference while preserving factual accuracy.
In terms of efficiency, \OurMethod\ also achieves the fastest editing speed. For instance, editing 100 examples on zsRE takes only \textbf{8.85s}, compared to 9.14s for MALMEN.
Such results underscore the practicality of our method for real-world use, where both speed and scalability are crucial.  
Figure~\ref{num_batch} further validates the robustness of \OurMethod\ by illustrating its stable performance under varying numbers of edits, showing that it can handle both small- and large-scale editing tasks without significant degradation. 
Although \OurMethod\ is not always the best on every single metric, its consistent overall superiority across models, metrics, and editing scales highlights its effectiveness, versatility, and strong potential for deployment in diverse real-world multi-scenario applications.

\subsection{Ablation Study}
To highlight the role of each component, Table~\ref{ablation} presents the ablation results for \OurMethod, where the update coefficient for each token is defined as \(c_i = -\eta \sum_{j} K_{ij} \tilde{K}_{ij}\).  
Removing the hierarchical orthogonal spread leads to clear drops, particularly in Specificity, confirming the necessity of spreading updates across layers.  
Excluding the update coefficient \(c_i\) almost eliminates editing ability, while removing it only from the regularization term also causes severe degradation, showing its importance for balancing factual alignment and knowledge preservation.  
Finally, removing network training results in sharp declines across all metrics. This demonstrates that the hypernetwork is indispensable, and it further highlights the strength of our residual design: by explicitly modeling \(R\), our method not only enables direct parameter updates but also provides structured signals that make the hypernetwork better aligned with editing objectives.

\begin{table}[!htbp]
  \centering
  \small
  \renewcommand{\arraystretch}{1.2}

  \resizebox{\linewidth}{!}{%
  \begin{tabular}{lccc}
    \toprule
    \textbf{Variant} & \textbf{Efficacy} & \textbf{Generalization} & \textbf{Specificity} \\
    \midrule
    \rowcolor{softpeach}
    \textbf{HORSE} & \textbf{90.25} & \textbf{86.27} & \textbf{60.24} \\
    w/o orthogonal spread
      & 88.72{\color{blue}\scriptsize{$\downarrow$ 1.53}}
      & 85.76{\color{blue}\scriptsize{$\downarrow$ 0.51}}
      & 58.03{\color{blue}\scriptsize{$\downarrow$ 2.21}} \\
    w/o $c_i$
      & 0.05{\color{blue}\scriptsize{$\downarrow$ 90.20}}
      & 0.05{\color{blue}\scriptsize{$\downarrow$ 86.22}}
      & 0.00{\color{blue}\scriptsize{$\downarrow$ 60.24}} \\
    w/o $c_i$ in $L_{\text{\OurMethod}}$
      & 22.84{\color{blue}\scriptsize{$\downarrow$ 67.41}}
      & 22.05{\color{blue}\scriptsize{$\downarrow$ 64.22}}
      & 16.89{\color{blue}\scriptsize{$\downarrow$ 43.35}} \\
    w/o network trainging
      & 37.24{\color{blue}\scriptsize{$\downarrow$ 53.01}}
      & 36.15{\color{blue}\scriptsize{$\downarrow$ 50.12}}
      & 39.40{\color{blue}\scriptsize{$\downarrow$ 20.84}} \\
    \bottomrule
  \end{tabular}}
  \caption{Ablation study of \OurMethod.}
  \label{ablation}
\end{table}

\subsection{Post-edited Model Analysis}

We assess post-editing performance by applying different methods to 8,000 zsRE edits on LLaMA2 and evaluating across several widely used benchmarks such as MMLU~\cite{mmlu}, NLI, and CommonQA, thereby examining potential side effects on the original model’s general abilities.
As shown in Table~\ref{post}, after massive editing, \OurMethod\ exhibits smaller overall side effects than competing approaches.

\begin{table}[!htbp]
\centering
\setlength{\tabcolsep}{3.5pt}
\renewcommand{\arraystretch}{1.1}
\begin{adjustbox}{max width=\linewidth}
\begin{tabular}{lcccccc}
\toprule
\textbf{Method} & \textbf{MMLU} & \textbf{NLI} & \textbf{MRPC} & \textbf{QNLI} & \textbf{QQP} & \textbf{CommonQA} \\
\midrule
\rowcolor{softblue}
Pre-edited & 13.93 & 50.21 & 81.52 & 49.92 & 53.35 & 32.84 \\
MEMIT    & \textbf{21.83} & 43.12 & \textbf{81.23} & 49.46 & \underline{53.83} & 20.23 \\
MEND     & 10.02 & 33.54 &  0.00 & 50.54 &  0.00 & 19.74 \\
PMET     & 14.17 & 46.13 & 42.52 & 49.59 & 53.47 & 25.63 \\
MALMEN   & 13.39 & \underline{47.48} & 34.66 & 51.03 & 27.38 & \underline{29.81} \\
\rowcolor{softpeach}
\textbf{\OurMethod}     & \underline{20.27} & \textbf{50.18} & \underline{80.31} & \textbf{53.25} & \textbf{54.06} & \textbf{29.87} \\
\bottomrule
\end{tabular}
\end{adjustbox}
\caption{Evaluation on the post-edited model}
\label{post}
\end{table}

\section{Conclusion}
In this work, we present \OurMethod, which achieves state-of-the-art performance in massive model editing. 
Extensive experiments demonstrate consistent improvements over prior methods, with an average gain of 6.26\% and a 10.12\% boost in specificity. 
These results highlight the effectiveness and robustness of our approach, enabling faster, more precise, and stable large-scale edits while preserving the model’s core capabilities. 
Beyond its strong performance, our approach also highlights the underexplored role of residual propagation across transformer layers, opening new directions for advancing model editing toward more controllable and trustworthy LLMs.

\clearpage

\bibliographystyle{IEEEbib}
\bibliography{strings,refs}
\end{document}